\documentclass{article}
\usepackage{spconf,amsmath,graphicx,booktabs,multirow,color}
\usepackage{amssymb}
\title{CS-R-FCN: Cross-supervised Learning for Large-scale Object Detection}
\name{Ye~Guo\qquad Yali~Li${^{*}}$\qquad Shengjin~Wang\thanks{This work was supported by the National Natural Science Foundation of China under Grant Nos.61701277 and 61771288 and the state key development program in 13th Five-Year under Grant No. 2017YFC0821601. *The corresponding author is Yali Li, E-mail: liyali13@tsinghua.edu.cn}}
\address{Beijing National Research Center for Information Science and Technology,
\\Department of Electronic Engineering, Tsinghua University, Beijing 100084, China}
\begin{document}
\topmargin=0mm
\maketitle
\begin{abstract}
Generic object detection is one of the most fundamental problems in computer vision, yet it is difficult to provide all the bounding-box-level annotations aiming at large-scale object detection for thousands of categories. In this paper, we present a novel cross-supervised learning pipeline for large-scale object detection, denoted as CS-R-FCN. First, we propose to utilize the data flow of image-level annotated images in the fully-supervised two-stage object detection framework, leading to cross-supervised learning combining bounding-box-level annotated data and image-level annotated data. Second, we introduce a semantic aggregation strategy utilizing the relationships among the cross-supervised categories to reduce the unreasonable mutual inhibition effects during the feature learning. Experimental results show that the proposed CS-R-FCN improves the mAP by a large margin compared to previous related works.
\end{abstract}


\section{Introduction}
\label{sec:introduction}

Object detection is a crucial task in computer vision, aiming to localize and recognize the objects in prescribed images or videos. Benefited from the application of deep convolution networks, object detection has made rapid development in recent years. Large-scale object detection is a challenging project to handle thousands of categories simultaneously in a unified framework, which is still not adequately addressed. The breakthrough of large-scale object detection will be of great significance to the further application of computer vision in reality. 

\begin{figure}[ht]
\centering
\centerline{\includegraphics[width=7.5cm]{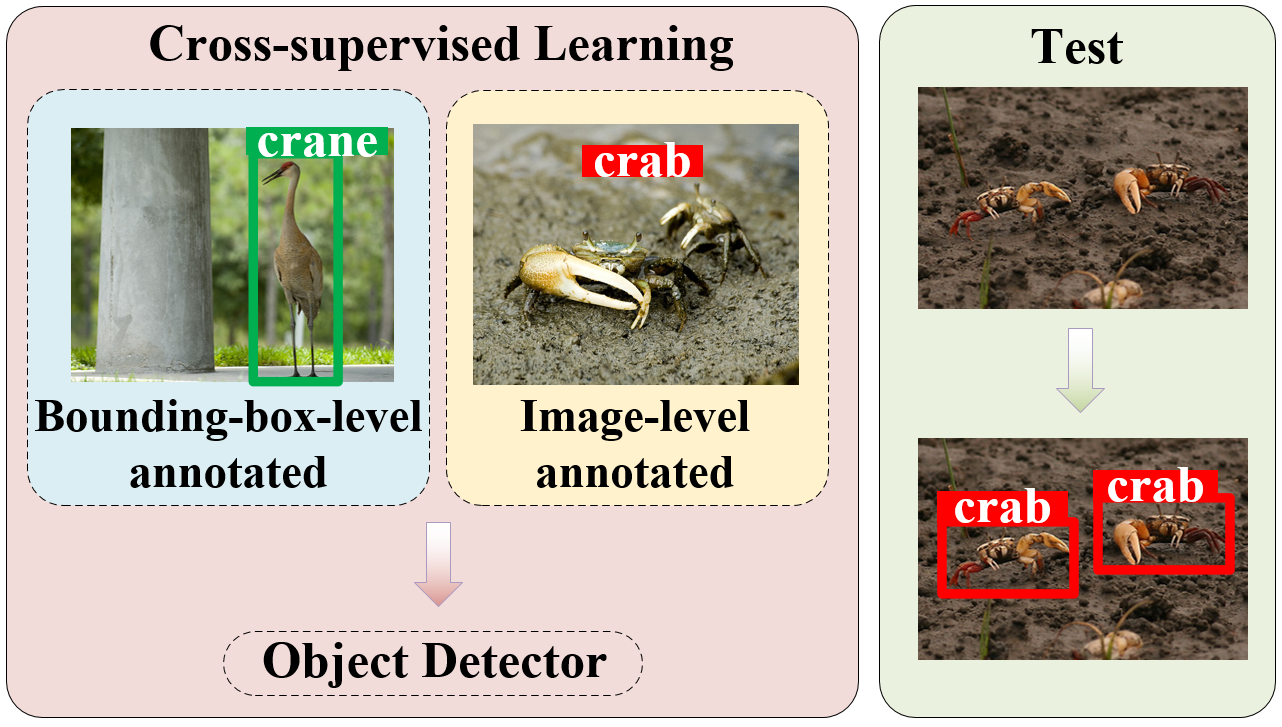}}
\vspace{-0.3cm}
\caption{By cross-supervised learning combining bounding-box-level annotated categories data and image-level annotated categories data, the proposed CS-R-FCN can localize and recognize the two-level annotated categories simultaneously.}
\label{fig:0}
\end{figure}

Object detection in terms of large-scale categories is now facing many difficult problems. First of all, labeling bounding boxes for enormous numbers of categories is so consuming that there are no sufficient samples supporting large-scale fully-supervised object detection training. It is necessary to develop a cross-supervised learning strategy combining the images data with bounding-box-level annotations as well as image-level annotations, as Fig.1 displays. Furthermore, the difficulty of feature learning increases rapidly due to the confusion among categories for cross-supervised large-scale object detection. It is vital to reorganize the supervision information and optimize the learning strategy.

\begin{figure*}[ht]
\centering
\centerline{\includegraphics[width=17cm]{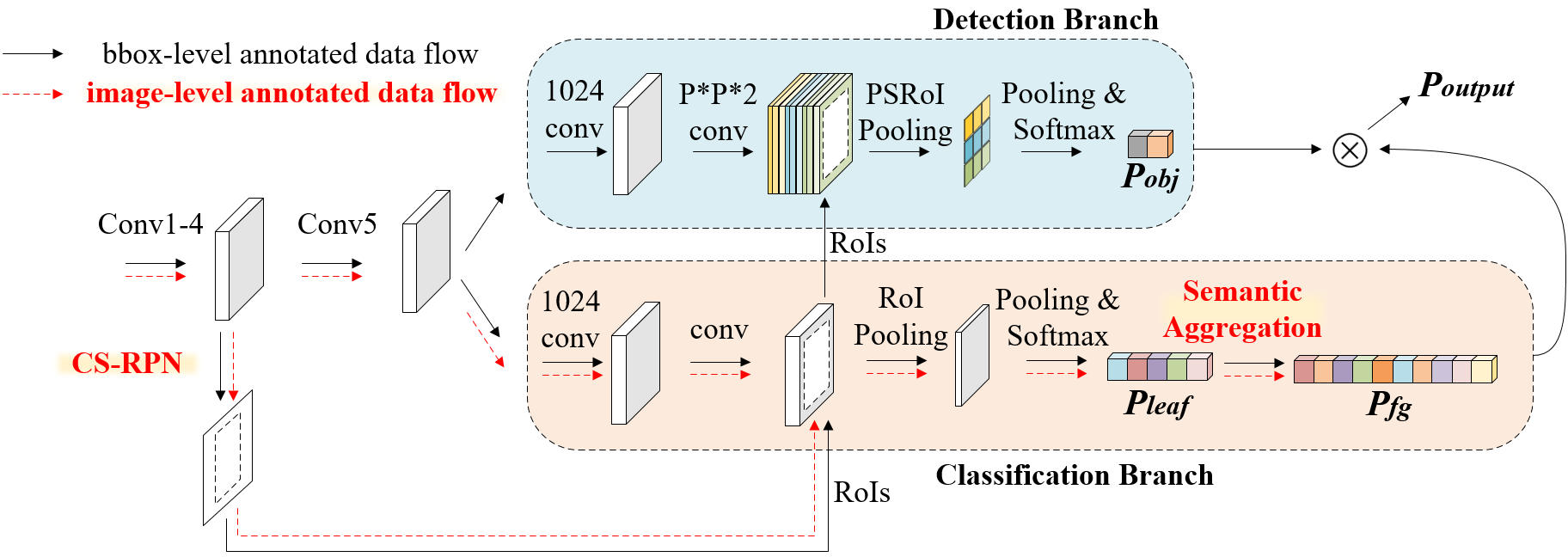}}
\vspace{-0.3cm}
\caption{Overview of the proposed CS-R-FCN. We propose to utilize the data flow of image-level annotated images in the RPN and classification branch, leading to cross-supervised learning in two-stage object detection. Moreover, we introduce the semantic aggregation strategy in the classification branch. Noteworthy, the regression process is omitted, due to the low correlation to this paper.}
\label{fig:1}
\end{figure*}

To deal with the partial lackness of bounding-box-level annotations, LSDA \cite{hoffman2014lsda,tang2018visual} adopts the method of domain adaptation and knowledge transfer, which trains the classifier for all categories and transfers the classifier to object detector. However, related solutions based on transfer learning cannot be trained end-to-end, short of elegance and efficiency. Based on YOLO v2 \cite{redmon2016you, redmon2017yolo9000}, YOLO-9000 proposes an end-to-end one-stage object detection pipeline to jointly train the image-level annotated and bounding-box-level annotated images simultaneously. To further boost the recognition accuracy, YOLO-9000 proposes to train the classifier hierarchically and perform multiple softmax operations over all co-hyponyms. However, limited by the one-stage pipeline, the final performance is still to be improved. Due to the superiority in the accuracy of two-stage object detection, such as Fast R-CNN  \cite{girshick2015fast}, Faster R-CNN \cite{ren2015faster}, R-FCN \cite{dai2016r} and R-FCN-3000 \cite{singh2018r}, it is valuable to realize more accurate cross-supervised learning based on the two-stage pipeline. Moreover, the hierarchical classification proposed by YOLO-9000 needs to add several intermediate categories nodes and extra space and time-consuming in the training and inference process, which can be further advanced.

In this paper, we propose a novel cross-supervised learning pipeline for large-scale object detection, denoted as \textbf{CS-R-FCN}. Our main contributions are as follows. First, we propose to utilize the data flow of image-level annotated images in the fully-supervised two-stage object detector, leading to superior cross-supervised learning. Second, we introduce an efficient semantic aggregation strategy utilizing the relationships among cross-supervised categories to reduce the unreasonable mutual inhibition effects during the feature learning. Experimental results show that the proposed CS-R-FCN can obtain superior performance than previous related frameworks and achieve comparable accuracy with fully-supervised baseline models.

The remainder part of this paper is organized as follows. Section 2 introduces the proposed method for large-scale cross-supervised object detection. Section 3 states the details of our implementation and analyzes the experimental results and comparison with former works. In Section 4, our work is concluded.

\section{The Proposed Method}
\label{sec:method}

\subsection{Cross-supervised Learning}
\label{ssec:method_1}
Fig.2 shows the overall pipeline of the proposed CS-R-FCN. We adopt R-FCN-3000 \cite{singh2018r} as the fully-supervised baseline model, which is a two-stage object detection pipeline with appreciated efficiency and accuracy. In the baseline model, the head network is decoupled to the detection branch and classification branch. The detection branch is to obtain objectness scores and execute bounding box regression for each RoI, whose parameters are irrelevant to the categories. Meanwhile, the classification branch is trained for a classifier to confirm the specific category for each instance.

For cross-supervised learning, the bounding-box-level annotated data can be trained as ordinary detection implementation. However, on account of the lackness of bounding box labels, image-level annotated images cannot participate in the training of the RPN and detection branch, but have to be utilized to finetune the classification branch. Consequently, the most important issue for cross-supervised learning in a two-stage object detection pipeline is how to provide proposals with accurate labels for image-level labeled categories and jointly train the classification branch of the head network.

In the original pipeline, RPN \cite{ren2015faster} is trained regardless of categories. Empirically, giving images with distinct and similar objects of untrained categories, well-trained RPN can still generate reliable proposals correspondingly. Motivated by this, we add image-level annotated data flow in the RPN and propose a novel cross-supervised region proposal network pipeline, denoted as CS-RPN, as Fig.3 shows. Consistently with RPN, the CS-RPN is updated using bounding-box-level annotated data and generates proposals containing positive and negative samples correspondingly for the head network training. Different from RPN, in the process of cross-supervised training, image-level annotated images are loaded and proposals of high scores generated by CS-RPN are assigned positive samples of corresponding categories and utilized to the finetuning of the classification branch. However, the image-level annotated data do not contribute to the update of CS-RPN.

\begin{figure}[ht]
\centering
\centerline{\includegraphics[width=7.7cm]{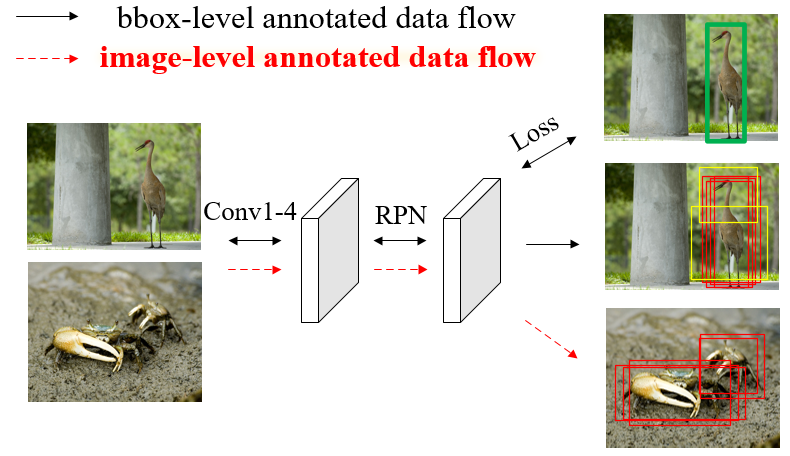}}
\vspace{-0.3cm}
\caption{Overview of the proposed CS-RPN. CS-RPN is updated using bounding-box-level annotations. Meanwhile, the image-level annotated data flow is added. The green, red, and yellow boxes represent the ground truth, positive samples, and negative samples correspondingly.}
\label{fig:2}
\end{figure}

Once the proposals and labels of these candidate boxes are acquired, it is practical to bind two kinds of data in the training batch and jointly train the classification branch. We denoted the loss function generated by bounding-box-level annotated data flow as $L_{b}$, and loss function generated by image-level annotated data flow as $L_{i}$. Namely, there are totally 5 parts in the overall loss function: anchor classification and regression loss in CS-RPN $L_{{CS-RPN}_{b}}$, smooth L1 loss for bounding box localization $L_{reg_{b}}$, objectness classification softmax cross-entropy loss $L_{obj_{b}}$, bounding-box-level annotated foregrounds classification loss $L_{cls_{b}}$ and image-level annotated foregrounds classification loss $L_{cls_{i}}$. As a result, the loss function of the cross-supervised object detection pipeline, denoted as $L_{cross}$ is:

\begin{equation}
    \begin{aligned}
    L_{cross} &= L_{b} + L_{i} \\
             &= L_{{CS-RPN}_{b}} + L_{reg_{b}} + L_{obj_{b}} + L_{cls_{b}} + L_{cls_{i}}
    \end{aligned}
\end{equation}

\subsection{Semantic Aggregation}
\label{ssec:method_3}
Suppose the predicted scores are $\left [ s_1, s_2,...s_C \right ]$, where $C$ is the number of target categories. Through softmax operation, the probability output for each category is $p_i={e}^{s_i}/\sum _{j=1}^{C}{e}^{s_j}$. Applying the softmax-based cross-entropy loss function, there are mutual inhibition effects in the feature learning among the training samples belonging to different categories. Faced with large-scale object detection task for tens of thousands of target categories, the dataset is so enormous that the relationships among target categories become complex. Based on the thesis stated above, if \textit{A} is a leaf category, and \textit{B} is a ancestor category of \textit{A} semantically, it is inappropriate to execute softmax-based classification on \textit{A} and \textit{B} simultaneously, which forms a unreasonable competition.

\begin{figure}[ht]
\centering
\centerline{\includegraphics[width=7.8cm]{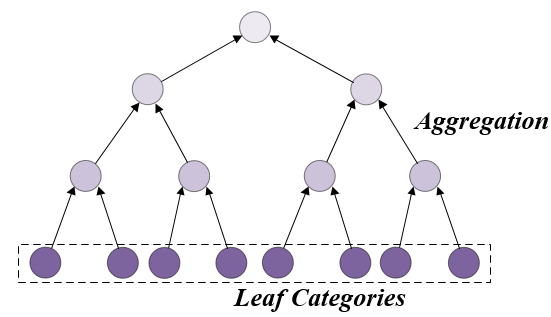}}
\vspace{-0.3cm}
\caption{All the cross-supervised categories are arranged in a unified tree depending on WordNet, and the leaf nodes are selected to be classified in the training. In inference, the probabilities of ancestor categories are aggregated by the probabilities of its descendant leaf categories.}
\label{fig:3}
\end{figure}

Motivated by this, we build a hierarchical structure to model the semantic relationships among the cross-supervised categories based on WordNet \cite{miller1995wordnet}, as Fig.4 shows. Further, the leaf nodes are selected, constituting a subset of cross-supervised categories denoted as $V_{leaf}$. Instead of categorizing all the cross-supervised categories, we just learn to classify on the leaf categories $V_{leaf}$. Suppose the predicted scores output from the classification branch for leaf categories are $\mathbf{s}=\left [ s_1, s_2,...s_{|V_{leaf}|} \right ]$, where $|V_{leaf}|$ is the total number of leaf categories. The softmax-based cross-entropy loss function is applied to train the novel classifier. In inference, the probabilities of ancestor categories are aggregated by summing all the probabilities of its descendant leaf categories ($\sigma\left({\cdot}\right)$), summarized as follows:

\begin{equation}
    P_{v_i}=\left \{
    \begin{aligned}
    &\left. {e}^{s_i} \middle/ \sum\nolimits_{j=1}^{|V_{leaf}|}{e}^{s_j} \right., & v_i\in{V_{leaf}}\\
    &\sum\nolimits_{{v_j}\in\sigma\left({v_i}\right)}P_{v_j}, & v_i\notin{V_{leaf}}
\end{aligned}
\right .
\end{equation}

By transferring the independent supervision of bounding-box-level annotated and image-level annotated categories to a unified and aggregated structure, the unreasonable conflicts in cross-supervised learning are significantly reduced. Noteworthy, due to the removal of ancestor categories samples, the training time for each epoch is further decreased. Compared to the hierarchical classification introduced by YOLO-9000 \cite{redmon2017yolo9000}, our proposed strategy is more efficient and easier to implement, performing the softmax operation for just one time.

\section{Experiment and Analysis}
\label{sec:exe}

\subsection{Datasets and Evaluation}
\label{sec:exe_1}
We combine bounding-box-level labeled \textit{ILSVRC DET train set} with image-level labeled \textit{ILSVRC CLS-LOC train set} \cite{deng2012ilsvrc}, forming a cross-supervised learning dataset. There are 285,356 bounding-box-level annotated images belonging to 544 categories and 648,937 image-level annotated images belonging to the other 506 categories in the cross-supervised learning dataset. 

For evaluation, we validate the proposed method on the \textit{ILSVRC DET val set}, which contains 200 categories and 20,121 images. We also evaluate the method on the \textit{ILSVRC CLS-LOC val hard set}, which contains 1,000 categories and 11,715 images selected from \textit{ILSVRC CLS-LOC val set} with more than one objects in each image, providing a more powerful validation for cross-supervised object detection task. We adopt the metrics from ILSVRC evaluation criteria \cite{deng2012ilsvrc}, mean Average Precision (mAP) at $IoU=0.5$.

\subsection{Implementation Details}
\label{sec:exe_2}
We adopt the same network implementation as the fully-supervised baseline model R-FCN-3000 \cite{singh2018r}. In the training and testing process, the images are resized to the resolution of $256\times256$. For cross-supervised learning, each training batch contains 16 bounding-box-level labeled images and 16 image-level labeled images, and horizontal flipping is used as a data augmentation technique. During training, warm-up learning is used for the first 1000 iterations and then the learning rate is increased to 0.015. The learning rate is dropped by a factor of 10 after 3 epochs. Totally, we train the network for 4 epochs on 2 GeForce GTX TITAN GPUs.

\subsection{Ablation Study}
\label{sec:exe_4}
It is noted that CS-RPN is pivotal in the proposed CS-R-FCN for providing reliable proposals for cross-supervised learning. To validate the effectiveness of CS-RPN, we present the AP and AR of the generated proposals for the object categories with bounding-box-level annotations and image-level annotations in Table 1. To be summarized, as the number of proposals increases, the accuracy of the generated proposals reduces and the recall improves. Without training, the CS-RPN can still generate reliable proposals for image-level annotated categories. When we set the proposals number to 10, AP of bounding-box-level annotated categories and image-level annotated categories are 63.4\% and 47.6\% correspondingly. If the number of proposals is set to 300, AR of bounding-box-level annotated categories and image-level annotated categories are 81.0\% and 75.1\% correspondingly. Therefore, in order to guarantee the accuracy in the cross-supervised learning, the proposals number for each image-level annotated image is set to 10. Meanwhile, the proposals number is set to 300 for inference to detect the targets completely.

\subsection{Comparison with state-of-the-art}
\label{sec:exe_3}
Table 2 summarizes the comparison of the training data and overall detection performance with previous related works. The proposed CS-R-FCN is compared with existing related methods such as LSDA, YOLO-9000, and R-FCN-3000 on the same \textit{ILSVRC DET val set}. Without semantic aggregation, the proposed CS-R-FCN achieves the mAP of 31.6\%. By applying the semantic aggregation, 39 ancestor categories and 14,013 training samples are filtered in the cross-supervised training, and the mAP of CS-R-FCN is further improved to 36.2\%.

Compared with LSDA and LSDA-VSKT, which also conduct cross-supervised learning with 100 bounding-box-level annotated categories and 100 image-level annotated categories, the proposed CS-R-FCN improve the mAP by a large margin of 14.1\% and 9.3\% correspondingly. Compared with YOLO-9000, which is cross-supervised trained combining bounding-box-level annotated MS COCO \cite{lin2014microsoft} with 80 categories and image-level annotated ImageNet release dataset with 9000 categories, the proposed CS-R-FCN boost the mAP significantly by 16.5\%. Compared to the fully-supervised baseline model R-FCN-3000, we obtain comparable performance with less bounding-box-level annotated data. Moreover, we obtain the mAP of 31.9\% among the
506 image-level annotated categories on \textit{ILSVRC CLS-LOC
val hard set}, further verifying the effectiveness of the proposed cross-supervised
learning pipeline.

\begin{table}[t]\small
\label{table1}
\centering
\vspace{0.2cm}
\setlength{\tabcolsep}{1.1mm}{
\begin{tabular}{ c | c | c | c | c | c | c}
\toprule[1.5pt]
No. of Proposals&10&20&50&100&200&300 \\ \midrule[1pt]
{AP (all)}&\textbf{55.8}&49.0&34.7&19.7&10.9&7.6 \\
{AP (bbox-level)}&\textbf{63.4}&54.1&39.8&23.0&12.4&8.4 \\
{AP (image-level)}&\textbf{47.6}&43.5&30.0&17.0&9.6&6.8 \\ \midrule[1pt]
{AR (all)}&34.5&39.3&53.6&66.0&74.1&\textbf{77.7}\\
{AR (bbox-level)}&40.0&46.6&60.2&71.3&78.2&\textbf{81.0} \\
{AR (image-level)}&26.6&33.7&48.6&61.8&70.9&\textbf{75.1} \\ \bottomrule[1.5pt]
\end{tabular}}
\caption{Average precision (AP) and average recall (AR) for proposals generation of CS-RPN on \textit{ILSVRC CLS-LOC val hard set}.}
\end{table}

\begin{table}[t]\small
\label{table2}
\centering
\vspace{0.2cm}
\setlength{\tabcolsep}{1.1mm}{
\begin{tabular}{ c | c | c | c }
\toprule[1.5pt]
{Method}&{bbox-level}&{image-level}&{mAP}\\ \midrule[1pt]
{LSDA \cite{hoffman2014lsda}}&100&100&22.1\\
{LSDA-VSKT \cite{tang2018visual}}&100&100&26.9\\
{YOLO-9000 \cite{redmon2017yolo9000}}&80&9,000&19.7\\
{R-FCN-3000 \cite{singh2018r}}&1,000&0&36.0\\ \midrule[1pt]
{CS-R-FCN w/o SA}&544&506&31.6 \\
{CS-R-FCN w SA}&505&506&\textbf{36.2}\\ \bottomrule[1.5pt]
\end{tabular}}
\caption{Comparison with previous works on \textit{ILSVRC DET val set}. SA denotes semantic aggregation.}
\end{table}

\section{Conclusion}
\label{sec:conclusion}
In this paper, we present a novel cross-supervised learning pipeline CS-R-FCN for large-scale object detection. First of all, we propose to utilize the data flow of image-level annotated images in the RPN and head network, leading to cross-supervised training in a more accurate two-stage object detector. Besides, we design an efficient strategy of semantic aggregation to decrease the unreasonable competition in cross-supervised learning. Experimental results show that the proposed CS-R-FCN can outperform previous related works and obtain comparable performance with the fully-supervised baseline model. 

\bibliographystyle{IEEEbib}
\bibliography{strings,refs}

\end{document}